\documentclass[10pt,twocolumn,letterpaper]{article}

\usepackage{iccv}
\usepackage{times}
\usepackage{epsfig}
\usepackage{graphicx}
\usepackage{amsmath}
\usepackage{amssymb}
\usepackage{color}

\AtBeginDocument{
}

\usepackage{graphicx, amsmath, amssymb, caption, subcaption, multirow, overpic, textpos}
\usepackage[table]{xcolor}
\usepackage[british, american]{babel}
\usepackage[pagebackref=true, breaklinks=true, letterpaper=true, colorlinks, citecolor=citecolor, linkcolor=linkcolor, bookmarks=false]{hyperref}
\definecolor{citecolor}{HTML}{0071BC}
\definecolor{linkcolor}{HTML}{ED1C24}
\usepackage{wrapfig}

\newlength\savewidth\newcommand\shline{\noalign{\global\savewidth\arrayrulewidth
  \global\arrayrulewidth 1pt}\hline\noalign{\global\arrayrulewidth\savewidth}}
\newcommand{\tablestyle}[2]{\setlength{\tabcolsep}{#1}\renewcommand{\arraystretch}{#2}\centering\footnotesize}
\renewcommand{\paragraph}[1]{\vspace{1mm}\noindent\textbf{#1}}

\newcolumntype{x}[1]{>{\centering\arraybackslash}p{#1pt}}
\newcolumntype{y}[1]{>{\raggedright\arraybackslash}p{#1pt}}
\newcolumntype{z}[1]{>{\raggedleft\arraybackslash}p{#1pt}}

\newcommand{\app}{\raise.17ex\hbox{$\scriptstyle\sim$}}

\definecolor{deemph}{gray}{0.6}

\definecolor{baselinecolor}{gray}{.9}
\newcommand{\baseline}[1]{\cellcolor{baselinecolor}{#1}}

\usepackage[ruled,vlined]{algorithm2e}
\usepackage{listings}
\usepackage{multicol,multirow}
\usepackage{booktabs}

\usepackage{arydshln}
\usepackage{blindtext}
\newcommand{\ourmethod}{{MFF }}

\newcommand{\more}[1]{\footnotesize {\textcolor[RGB]{57,181,74}{#1}}}

\newcommand{\syadd}[1]{\textcolor{red}{#1}}
\newcommand{\green}[1]{\textcolor{green}{#1}}
\newcommand{\red}[1]{\textcolor{red}{#1}}

\newcommand{\comment}[1]{{}}
\newcommand{\default}[1]{\baseline{\textbf{#1}}}

\usepackage{enumitem}
\setlist[itemize]{noitemsep,leftmargin=*,topsep=0in}
\setlist[enumerate]{noitemsep,leftmargin=*,topsep=0in}

\usepackage[pagebackref=true,breaklinks=true,letterpaper=true,colorlinks,bookmarks=false]{hyperref}

\iccvfinalcopy % *** Uncomment this line for the final submission

\def\iccvPaperID{1155} % *** Enter the ICCV Paper ID here

\ificcvfinal\fi
\begin{document}

%%%%%%%%% TITLE
\title{Improving Pixel-based MIM by Reducing Wasted Modeling Capability}

\author{
  \hspace{-0.25cm}\textbf{Yuan Liu$^{1}$, Songyang Zhang$^{1,\ddagger}$, Jiacheng Chen$^2$, Zhaohui Yu$^3$, Kai Chen$^{1,\ddagger}$} \\
  \hspace{-0.25cm}\textbf{Dahua Lin$^{1,3}$} \\ 
 \hspace{-0.25cm}$^1$Shanghai AI Laboratory \quad $^2$Simon Fraser University \quad $^3$The Chinese University of Hong Kong \\
 \normalsize{\hspace{-1em}$^{\ddagger}$ Corresponding author} \\
  }

\maketitle
\ificcvfinal\thispagestyle{empty}\fi

\begin{abstract}
  There has been significant progress in Masked Image Modeling (MIM). Existing MIM methods can be broadly categorized into two groups based on the reconstruction target: pixel-based and tokenizer-based approaches. The former offers a simpler pipeline and lower computational cost, but it is known to be biased toward high-frequency details. In this paper, we provide a set of empirical studies to confirm this limitation of pixel-based MIM and propose a new method that explicitly utilizes low-level features from shallow layers to aid pixel reconstruction. By incorporating this design into our base method, MAE, we reduce the wasted modeling capability of pixel-based MIM, improving its convergence and achieving non-trivial improvements across various downstream tasks. To the best of our knowledge, we are the first to systematically investigate multi-level feature fusion for isotropic architectures like the standard Vision Transformer (ViT). Notably, when applied to a smaller model (e.g., ViT-S), our method yields significant performance gains, such as 1.2\% on fine-tuning, 2.8\% on linear probing, and 2.6\% on semantic segmentation. Code and models are available in MMPretrain\footnote{\url{https://github.com/open-mmlab/mmpretrain}}.
\end{abstract}
\section{Introduction}
\label{sec:intro}
Self-supervised learning (SSL) has made remarkable progress in language and computer vision. Masked Image Modeling (MIM) is an effective framework in image SSL, which boasts a simple training pipeline, a few handcrafted data augmentations, and high performance across downstream tasks. In the pioneering work of BEiT~\cite{BEiT}, 40\% of the input image is masked, and the model is trained to capture the semantics of the masked patches by reconstructing the DALL-E~\cite{DALLE} output features. To simplify pre-training and reduce computational overhead, MAE~\cite{MAE} only feeds the visible tokens into the encoder and encourages the decoder to reconstruct the raw pixels of masked patches. More recently, follow-up works have focused on adding auxiliary tasks or using large-scale pre-trained models to produce reconstruction targets. For instance, CMAE~\cite{CMAE} explicitly adds a contrastive task and optimizes it in conjunction with the MIM task, while MILAN~\cite{MILAN} and BEiT-v2~\cite{BEiTv2} employ multimodal pre-trained models such as CLIP~\cite{CLIP} to generate the reconstruction features. 
\begin{figure}[t!]
\centering
\includegraphics[width=\linewidth]{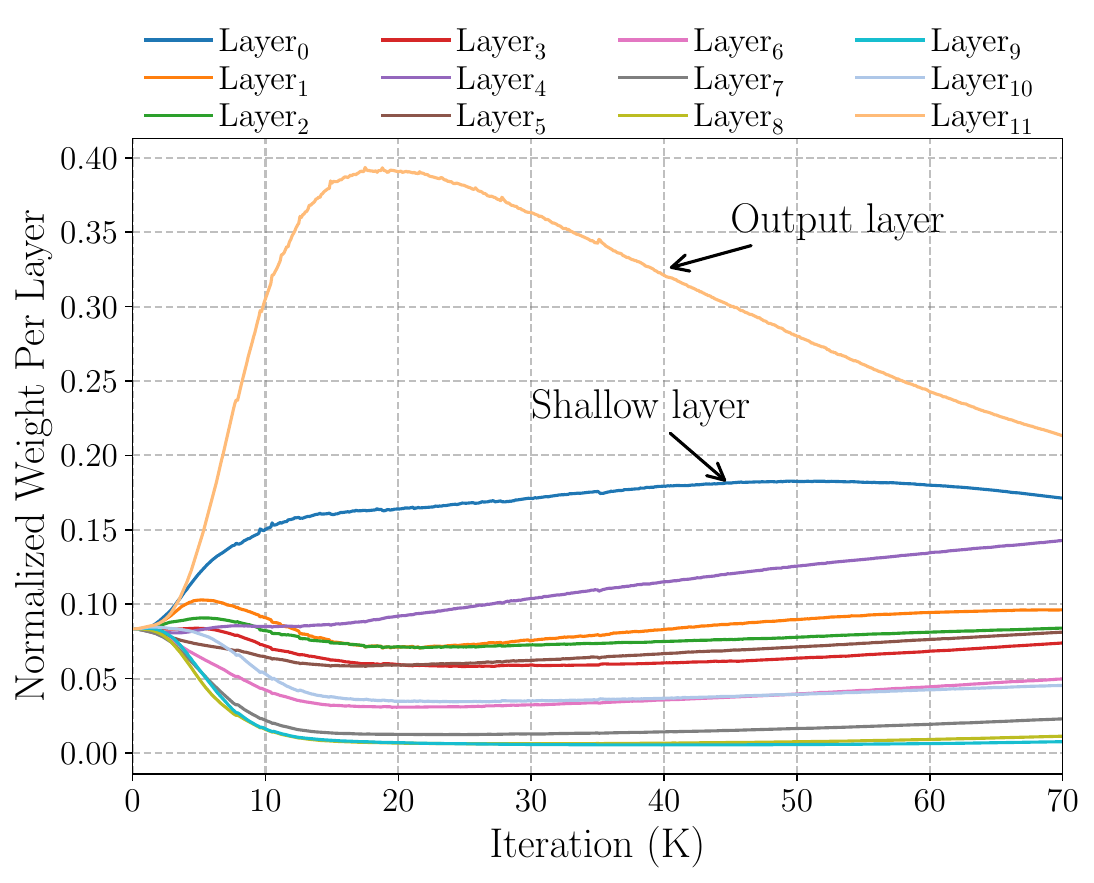}
\vspace{-2em}
\caption{\textbf{Fusing shallow encoder layers for MAE.} During the training process, MAE increasingly relies on shallow layers for the pixel reconstruction task, demonstrating a bias toward low-level features.}
\label{fig:teaser}
\end{figure}

Among the various MIM methods available, pixel-based approaches such as MAE \cite{MAE} are particularly interesting because of their simple pre-training pipeline and minimal computational overhead. However, these methods are typically biased towards capturing high-frequency details due to their emphasis on reconstructing raw pixels \cite{BEiT, pixmim}. As a result, they waste a significant amount of modeling capability that could be better utilized to capture low-frequency semantics. Our objective is to reduce this waste of modeling capacity, aiming for an improved quality of learned representation for downstream visual tasks. Toward this goal, we design two pilot experiments based on the representative work of MAE \cite{MAE} to uncover its neglected design aspects.

\begin{figure}[t!]
\centering
\includegraphics[width=\linewidth,scale=1.00]{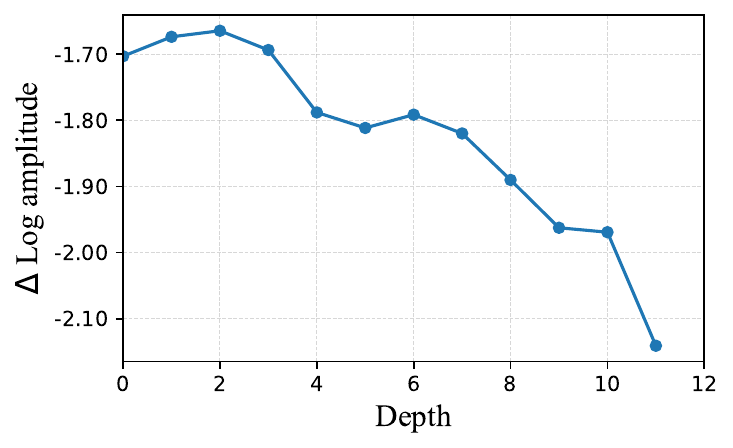}
\caption{\textbf{Frequency analysis of MAE.} Higher log amplitude denotes more high-frequency information. Shallow layers contain more high-frequency information (or low-level information) than deep layers.}
\label{fig:freq_ana}
\end{figure}

\begin{enumerate}[label={\bf {{(\arabic*)}}},leftmargin=*,topsep=0.5ex,itemsep=0ex,partopsep=0.75ex,parsep=0.75ex,partopsep=0pt,wide, labelwidth=0pt,labelindent=0pt]
    \comment{\item \textbf{Reconstruction Shortcut}: As the learning objective of MAE is to recover the original images, introducing the low-level features as the shortcut into the decoding process is a straightforward idea. We first examine the influence of utilizing multi-level features for reconstruction quantitatively. Specifically, we apply a parametric weighted average to features of all layers for decoding in MAE and plot the curve of the weight of multi-level features during the model optimization in \autoref{fig:teaser}.
    We found that the model increasingly relies on the features of these shallow layers, such as the first layer, throughout the entire training process.}

    \item \textbf{Fuse Shallow Layers}: Rather than solely using the output layer for pixel reconstruction, we implement a weight-average strategy to fuse the output layer with all previous layers. The weights assigned to each layer are normalized and dynamically updated during the pre-training process, with their absolute values indicating the significance of each layer for the reconstruction task. We track the changes in these weights and illustrate them in \autoref{fig:teaser}. As depicted in the figure, the model increasingly relies on the features of shallow layers as training progresses.
    
    \item \textbf{Frequency Analysis}: To further understand the property of the representation learned with MAE, we analyze the frequency response of each layer's feature. We adopt the tools proposed by \cite{howdo} to transform the encoder features into the frequency domain and visualize the relative log amplitude of the transformed representation in \autoref{fig:freq_ana}. Typically, a higher amplitude indicates that the feature produced by one layer contains \textit{more high-frequency information}. We empirically find that the shallow layers contain significantly more high-frequency components than deep layers, which are mostly related to low-level details (\eg, textures).

    \comment{\item \textbf{Frequency Analysis}: We further analyze the frequency response of each layer in the MAE model. By transforming the representation of each layer into the frequency domain, we calculate the relative log amplitude of the Fourier-transformed representation, similar to the approach in \cite{howdo}. A higher amplitude indicates that a layer contains more high-frequency information. Empirically, we categorize layers above the 6th layer as deep layers and those below the 6th layer as shallow layers. As shown in \autoref{fig:freq_ana}, compared to deep layers, shallow layers contain significantly more high-frequency components, which are mostly related to low-level details (\eg, textures).}
    
\end{enumerate}

\comment{Based on the two experiments described above, it is evident that the raw pixel reconstruction task causes a bias toward low-level details, which may be a unique drawback of these pixel-based MIM methods (not exist in these MIM approaches, using another tokenizer). Furthermore, the low accuracy of linear probing directly reflects the limitations of using raw image as target since linear separability necessitates that the extracted features be semantically distinct from each other. To compensate for this drawback, we explicitly incorporate low-level features into the output layer by reusing the \textbf{M}ulti-level \textbf{F}eature \textbf{F}usion (\red{\ourmethod}) strategy employed in the first pilot experiment. This strategy is straightforward and intuitive and can be easily integrated into most existing pixel-based MIM approaches without incurring significant computational resources. Despite its simplicity, it has the following two implications:}

\comment{\syadd{Based on the results of the pilot experiments, it is evident that the representation learned in MAE is high-frequency biased and find that the raw pixel reconstruction task aggressively requires the low-level feature when introducing shortcuts. Moreover, the biased representation  causes the poor linear separability in MAE(such as the low accuracy of linear probing). To compensate the limitation of pixel-based MIM, we propose a simple yet effective strategy by incorporating the low-features for pixel reconstruction in this work.}}

Based on our analysis of the pilot experiments, we can conclude that the pixel reconstruction task exhibits a bias towards low-level details. This is evident from the low linear probing accuracy, which highlights the constraints of the pixel reconstruction task. Namely, it requires features that are semantically distinct enough to achieve linear separability. To address this limitation, we propose to explicitly incorporate low-level features obtained from shallow layers into the output layer for the pixel reconstruction task. By doing so, we alleviate the burden of the model having to focus excessively on these low-level details, allowing it to spend its modeling abilities to capture these high-level semantics.

We denote the proposed method as \textbf{M}ulti-level \textbf{F}eature \textbf{F}usion (\red{\ourmethod}). Specifically, we extend the usage of the fusion strategy in the first pilot experiment and systematically investigate the design choices of multi-feature fusion, such as feature selection and fusion strategies. Despite the simplicity of the proposed method, it is a drop-in solution for unleashing the full modeling potential of pixel-based MIM approaches and has the following advantages:

\begin{enumerate}[label={\bf {{(\arabic*)}}},leftmargin=*,topsep=0.5ex,itemsep=-0.5ex,partopsep=0.75ex,parsep=0.75ex,partopsep=0pt,wide, labelwidth=0pt,labelindent=0pt]
    \item Employing multi-level feature fusion can enhance the training efficiency of MAE by approximately $\sim$5x, thus helping to reduce the carbon footprint. For example, by pre-training MAE with this strategy for only 300 epochs, we achieve semantic segmentation results that are on par with those obtained after 1600 epochs in the original paper. 
    \item We also consistently and significantly improve performance across all downstream tasks, including semi-supervised fine-tuning and linear probing. Notably, with a small model such as ViT-S, we outperform MAE by 2.8\% on linear probing, 2.6\% on semantic segmentation, and 1.2\% on fine-tuning.
    \item After evaluating our model on four out-of-distribution datasets, we observe that the approach with multi-level feature fusion exhibits greater robustness than the base method.
\end{enumerate}

Furthermore, we conduct a thorough analysis to unveil how multi-feature fusion works for representation learning. Given the exploratory experiments from the perspective of latent features and optimization, we find that the fusion strategy attenuates high-frequency information in the latent features and flattens the loss landscapes. To summarize, our contributions are three-fold:

% \noindent To summarize, our contributions are three-fold:
\begin{itemize}[leftmargin=*,topsep=0pt,itemsep=0pt,noitemsep]
    \item Firstly, we develop a multi-level feature fusion strategy for isotropic backbones such as ViT, achieving superior results compared to various pixel-based MIM approaches.
    \item Secondly, we have conducted a thorough analysis of how this multi-level feature fusion strategy enhances the model from the perspectives of latent features and optimization. Our examination is meticulous and provides valuable insights.
    \item Lastly, we have performed extensive and rigorous ablation studies on the design details, which also strengthens the validity of our findings.
\end{itemize}

\section{Related Works}

\paragraph{Self-supervised Learning} reviously, many works\cite{ViT, DeiT} relied on abundant labeled datasets to achieve promising results. However, annotating this data requires a large number of human labors. Therefore, how to effectively capture useful semantics embedded in the abundance of data available on the Internet is currently a hot topic.
In recent times, self-supervised learning has witnessed tremendous growth in computer vision, following remarkable achievements in natural language processing. These methods cater to diverse inputs, including images\cite{simclr,BEiT,MAE,ibot}, videos\cite{2022MoQuad,hu2021contrast}, and multi-modality inputs\cite{CLIP,2021LearningTBP}. They capture rich semantic information by creating effective proxy tasks, such as contrastive learning and masked image modeling, in large amounts of unlabeled data. In comparison to supervised learning\cite{DeiT,DeiT-v2,DEiT-v3}, these self-supervised learning approaches have gradually outperformed them in numerous downstream tasks and possess immense potential to become the principal pre-training paradigm.

\paragraph{Feature Pyramid} Utilizing multi-level features has been extensively studied in previous years, and one of the most famous applications is the Feature Pyramid Network (FPN)\cite{FPN}. This technique has been widely used in many dense tasks such as object detection and semantic segmentation to improve the model's perception of objects of different scales. Incorporating FPN into existing designs in many works\cite{maskrcnn, upernet} has led to significant improvements. However, the multi-level feature fusion module only accepts features of different scales as input, limiting its adaptation to isotropic architectures such as ViT\cite{ViT}, in which features from different layers are of the same scale. In masked image modeling, most approaches choose ViT as their encoder due to the masked patch prediction task. Therefore, there are few works exploring multi-level feature fusion in this domain. Even though some works~\cite{ConvMAE,itpn} aim to explore multi-level fusion in masked image modeling, their applications are still limited to the traditional hierarchical architecture and do not address the issue of being biased toward low-level details for these pixel-based methods.

\section{Methods}
\begin{figure*}[!htbp]
\centering
\includegraphics[width=\linewidth,scale=1.00]{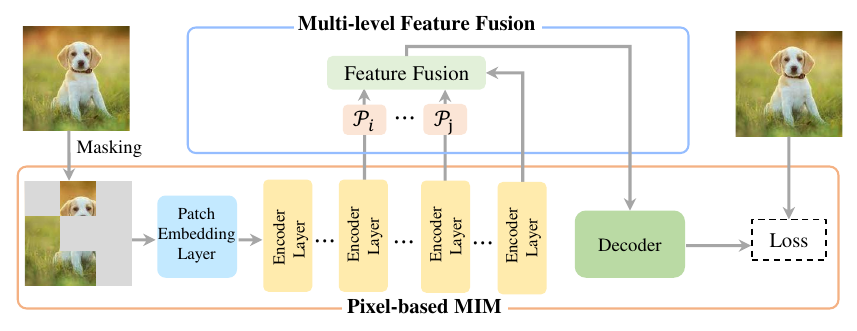}
\caption{\textbf{Multi-level feature fusion for pixel-based MIM.} The multi-level feature fusion (MFF) module can be inserted into existing pixel-based MIM approaches in a plug-and-play manner. $i, j \in \mathcal{W}$.}
\label{fig:main}
\end{figure*}

% Multi-level feature fusion can be incorporated into most of existing pixel-based MIM approaches without any efforts. 
We now introduce the proposed multi-level feature strategy for pixel-based MIM methods. In \autoref{method:intro2mim}, we first give a short revisit to the universal framework of pixel-based MIM approaches. Then \autoref{method:insert} describes how to insert multi-level feature fusion into pixel-based MIM approaches. Finally, two key components, the projection layer and fusion strategy, will be discussed in \autoref{method:proj_and_fusion}.

\subsection{Introduction to Pixel-based MIM}
\label{method:intro2mim}
We now introduce the unified formulation of recent pixel-based Masked Image Modeling(MIM) methods. This kind of MIM, using raw images as target, is a denoising autoencoder \cite{denoising}, and it follows a simple pipeline. Its primary objective is to predict raw pixel values of the original or post-processed images, such as the high-frequency filtered image in PixMIM \cite{pixmim}. When dealing with masked images, we can feed only visible tokens into the encoder or both visible and mask tokens. If only visible tokens are used for the encoder, both the mask token and the latent feature output by the encoder must be fed into the decoder. 
\comment{The \textbf{black} code shown in \autoref{alg: rgb-mim} represents the general framework for this approach.}

\subsection{MIM with Multi-level Feature Fusion}
\label{method:insert}
Multi-level feature fusion(MFF) can be incorporated into most existing pixel-based MIM approaches in a plug-and-play manner. \autoref{fig:main} gives an overview of the whole framework.
To keep the simplicity, we mainly focus on these steps relevant to MFF, leaving out other steps.  Given an image $\mathbf{I}\in \mathbb{R}^{H\times W\ \times3}$, we feed it into the encoder, $\mathbb{E}$, to get the latent representations:
\begin{equation}
    \label{eq:extract}
    \mathrm{X} = \mathbb{E}(\mathbf{I})
\end{equation}
The latent representations, denoted by $\mathrm{X} = \{x_0, x_1, ..., x_{N-2}, x_{N-1}\}$, correspond to the output feature from each transformer layer of the ViT, where $N$ represents the depth of the encoder. For the pilot experiment in \autoref{fig:teaser}, we fuse all-level features from each layer of the encoder. However, indiscriminately fusing all of them may introduce redundancy or even makes the model much harder to be optimized. But finding the most effective layers to fuse induces a large search space. To simplify the layer selection procedure, we follow the guidelines below:
\comment{To determine which level's and how many level's features should be used for fusion. We thoroughly investigate the layer selection strategy following those guidelines:}

\begin{enumerate}[label={\bf {{(\arabic*)}}},leftmargin=*,topsep=0.5ex,itemsep=-0.5ex,partopsep=0.75ex,parsep=0.75ex,partopsep=0pt,wide, labelwidth=0pt,labelindent=0pt]
    \item: We first conduct an ablation study to compare the results of fusing shallow layers or deep layers (as shown in  \autoref{fig:freq_ana}, shallow layers contain low-level features, and deep layers contain high-level features), and more details are presented in \autoref{sec:ablation}. The results show that utilizing the features of the shallow layers performed significantly better than deep ones. Thus intuitive analysis and the quantitative numbers both indicate that the shallow layer should be selected for fusion.
    \item: We then examine how many layers should be taken into consideration. In addition to the selected shallow layer and output layer, we also try to explore introducing different numbers of intermediate layers for fusion. We refer the reader to \autoref{sec:ablation} for more details of this experiment.
\end{enumerate}

We finally selected $M=5$ additional layers besides the last layer (6 layers in total) and the output features from those layers are used for fusion. We define the indices for these selected layers as $\mathcal{W}, |\mathcal{W}|=M$.
After that, we apply a projection layer, $\mathcal{P}_\text{i}$, to each of the additional $M$ layers before fusion. 
\begin{equation}
    \label{eq:proj}
    \Tilde{\mathrm{X}} = \{\mathcal{P}_\text{i}(x_i)\}_{i \in \mathcal{W}} + \{x_{N-1}\}
\end{equation}
Adding a projection layer to align the feature space between different levels' features is a common practice in self-supervised learning.

Finally, we introduce the fusion layer, $\mathcal{F}$,  to fuse multi-level features $\Tilde{\mathrm{X}}$:
\begin{equation}
    O = \mathcal{F}(\Tilde{\mathrm{X}})
\end{equation}
$O$ will be fed into the decoder for pixel reconstruction.

% \subsection{Projection and Fusion Layer}
\subsection{Instantiation of Projection and Fusion Layers}
\label{method:proj_and_fusion}
We further investigate the instantiation of the projection layer and the fusion layer. 

\noindent\textbf{Projection Layer}: In terms of the projection layer, indicated as $\mathcal{P}$, we focus on two popular options, namely linear projection and non-linear projection. Specifically, we instantiate the non-linear projection with the Linear-GELU-Linear structure. Our experiment has revealed that a simple linear layer is sufficient and effective within our framework. 

\noindent\textbf{Fusion Layer}: The fusion layer aims to gather low-level information from the features of shallower layers feature. We evaluate two commonly employed fusion methods: weighted average pooling and self-attention-based fusion. 

\begin{equation}
    \label{eq:pooling}
    O = \sum_{i \in \mathcal{W}}w_i\mathcal{P}_i(x_i) + w_{N-1}x_{N-1}
\end{equation}

 \noindent The weighted average pooling fusion is illustrated in \autoref{eq:pooling}. In this equation, $w_i$ refers to the weight assigned to each of the $M$ selected layers, while $w_{N-1}$ is assigned to the output layer. During the training process, all these weights are dynamically updated and summed up to 1. As for the self-attention method, we use an off-the-shelf transformer layer.

\begin{equation}
    \label{eq:attn}
    \hat{O} = \text{MultiHeadAttention}(([{\mathcal{P}_i(x_i)}_{i \in \mathcal{W}}, x_{N-1}])
\end{equation}

\noindent After the multi-head attention layer, we extract the transformed tokens corresponding to $x_{N-1}$ from $\hat{O}$ to use in pixel reconstruction. Experimental results demonstrate that the weighted average pooling strategy is comparable to self-attention for this purpose while also being simpler and more computationally efficient.

\hspace*{\fill}

Our method is intuitive and simple, and can be inserted into most of pixel-based MIM approaches without introducing non-negligible computational overhead. We evaluate it on MAE\cite{MAE} and PixMIM\cite{pixmim}, and more detailed results are shown in \autoref{sec:main_results}.
\section{Analysis}
In order to uncover the dark secret behind multi-level feature fusion, we first conduct a frequency analysis on the model (with and without MMF) in \autoref{sec:freq_ana}. In \autoref{sec:optim_ana}, we find multi-level feature fusion can be helpful to the optimization of the model, by flattening the loss landscape. Finally, in \autoref{sec:proprietary}, we apply the first pilot experiment introduced in \autoref{sec:intro} to EVA \cite{EVA} and supervised ViT \cite{MAE}, to investigate whether they too require low-level features and to confirm that the bias towards low-level details is the unique and inherent drawback of pixel-based MIM.
\label{sec:analysis}
\subsection{Frequency Analysis}
\label{sec:freq_ana}
\begin{figure}[t!]
\centering
\includegraphics[width=\linewidth,scale=1.00]{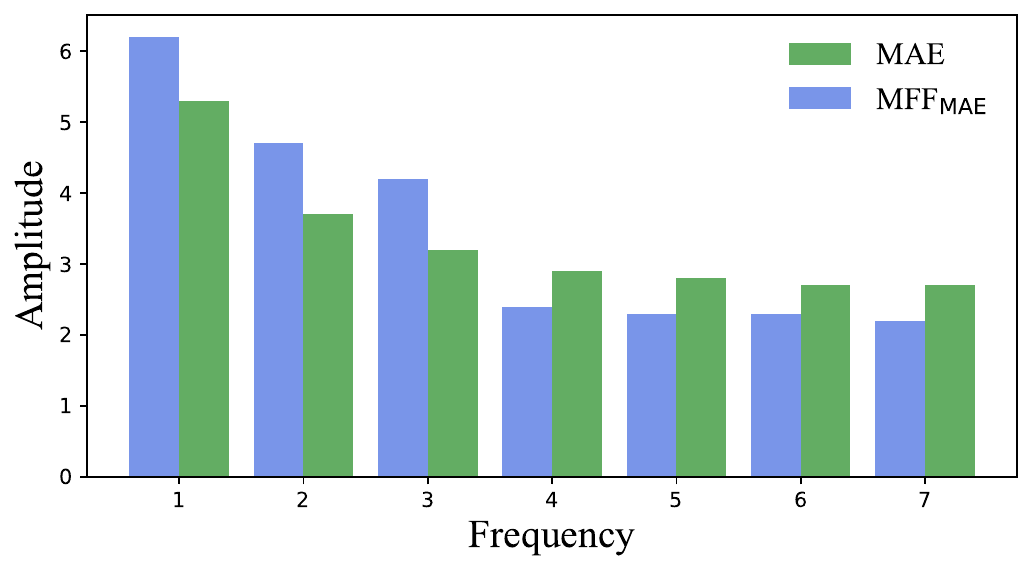}
\caption{\textbf{Frequency analysis of output layer features from the encoder.} Multi-level feature fusion reduces high-frequency components and enhances low-frequency components. The lowest-frequency band is labeled as \textbf{1}, while the high-frequency band is labeled as \textbf{7}.}
\label{fig:last_layer_freq}
\end{figure}

We employ multi-level feature fusion to enhance MAE \cite{MAE}, resulting in MFF$_\text{MAE}$. The aim of this fusion is to prevent the model from excessively focusing on low-level details. To investigate the change in frequency response before and after the fusion, we transform the feature from the last encoder layer into the frequency domain and calculate the amplitude of various frequency bands. As depicted in \autoref{fig:last_layer_freq}, multi-level feature fusion diminishes high-frequency responses and amplifies those that belong to the low-frequency range, which supports the efficacy of the fusion technique.

\subsection{Optimization Analysis}
\label{sec:optim_ana}
\begin{figure}[t!]
\centering
\includegraphics[width=\linewidth,scale=1.00]{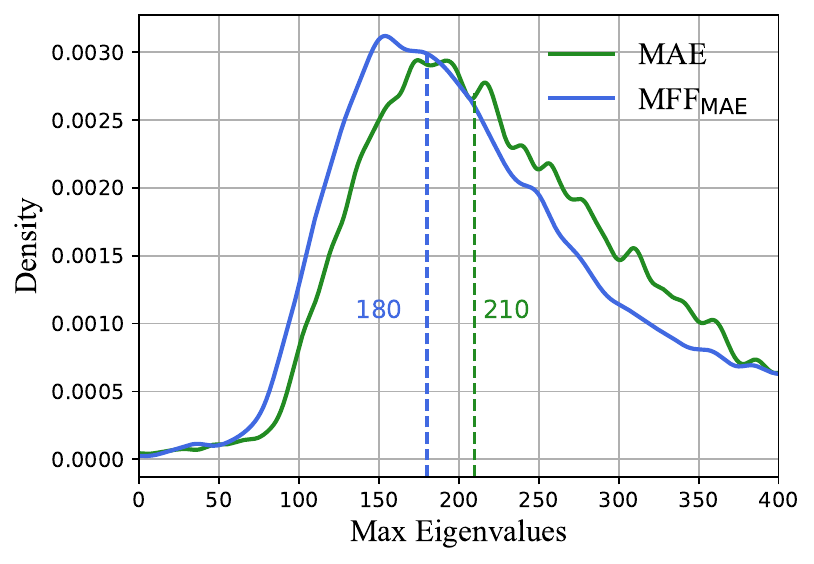}
\caption{\textbf{Hessian max eigenvalues spectrum.} Multi-level feature fusion reduces the magnitude of Hessian max eigenvalues to flatten the loss landscape.}
\label{fig:optim_eigen}
\end{figure}

Following \cite{howdo}, we analyze the Hessian max eigenvalue spectrum. Multi-level feature fusion has the additional benefit of reducing the magnitude of Hessian max eigenvalues. As shown in \autoref{fig:optim_eigen}, the expected hessian max eigenvalue of MFF$_\text{MAE}$ is smaller than that of MAE\cite{MAE}. Hessian max eigenvalues represent the local curvature of the reconstruction loss function, and this result suggests that multi-level feature fusion flattens the loss landscape. Large eigenvalues can impede neural network training~\cite{ghorbani2019investigation}. Therefore, \textit{multi-level feature fusion can help the model learn better representations by suppressing large Hessian eigenvalues.}

% \subsection{Biased toward low-level features only exists in pixel-based MIM}
\subsection{Feature Bias of Different Pre-training Methods}
\label{sec:proprietary}
\setlength{\abovecaptionskip}{5pt}
\begin{figure*}[htbp]
\centering
\includegraphics[width=\linewidth]{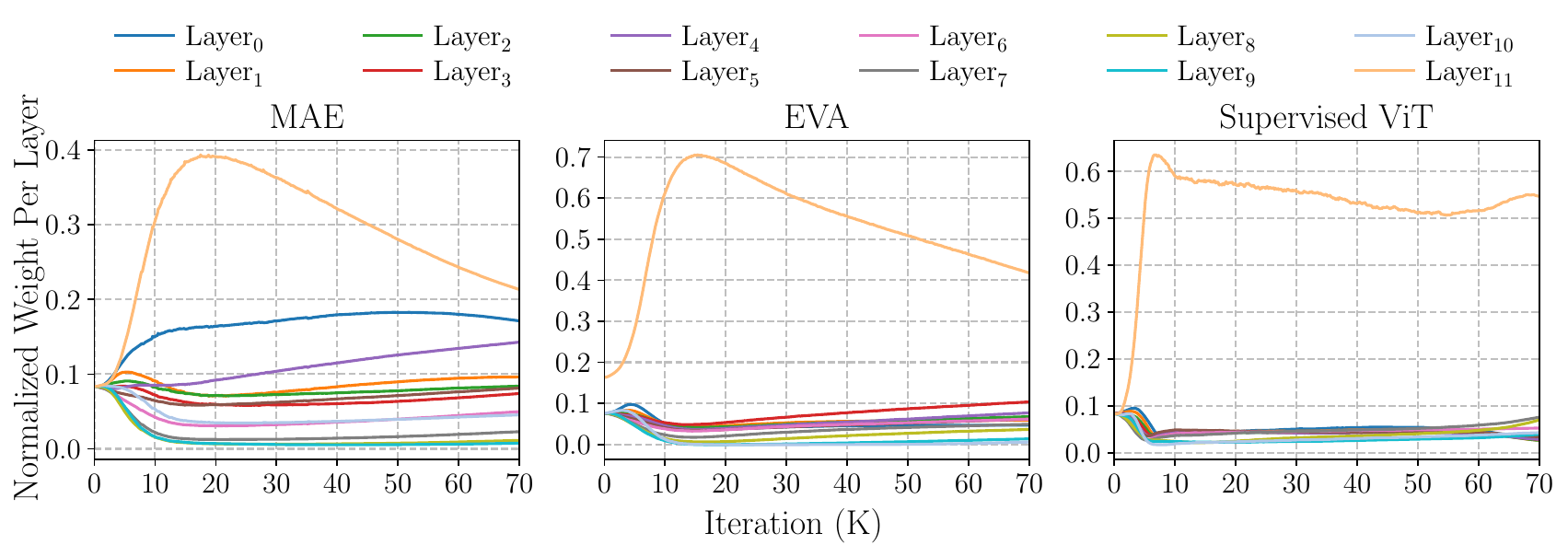}
% \vspace{-1.2em}
\caption{\textbf{MAE~\cite{MAE} attempts to extract low-level features from shallow layers, whereas EVA~\cite{EVA} and DEiT~\cite{DeiT} do not.} MAE refers to MIM approaches that use the raw pixel as the target, while EVA refers to MIM approaches that use additional tokenizers like CLIP\cite{CLIP}. Supervised ViT denotes supervised learning approaches. Compared to EVA and supervised ViT, MAE exhibits a greater affinity for low-level features and displays a more aggressive appetite for them.}
\label{fig:weight_vs_layer}
\end{figure*}
In order to investigate whether being biased towards low-level features is the sole and inherent drawback of pixel-based MIM, we introduce multi-level feature fusion to EVA\cite{EVA} and supervised ViT\cite{MAE}. EVA is one of the representative works that focuses on regressing high-level features produced by CLIP\cite{CLIP}, while supervised ViT is one of the works that require the model to map the input image to a semantic label. Both EVA and supervised ViT target high-level features that contain rich semantic information describing the input image. As shown in \autoref{fig:weight_vs_layer}, unlike MAE, the weight of the last layer's feature for both EVA and supervised ViT is significantly higher than that of the shallow layers. This observation suggests that the bias towards low-level features exhibited by these pixel-based MIM approaches is primarily caused by the raw-pixel reconstruction task.

\section{Experiment}
In~\autoref{sec:exp:settings}, we present the experimental settings for pre-training and evaluation. Next, in \autoref{sec:main_results}, we apply MFF to two MIM baselines, namely MAE~\cite{MAE} and PixMIM~\cite{pixmim}, and show the improvements brought by such design. In addition, we also evaluate the effectiveness of MFF using a smaller model (\eg, ViT-S) and fine-tune the pre-trained model under a low-shot setting. To evaluate the robustness of our proposed method, \autoref{sec:robust_eval} includes additional analyses that assess the robustness of pre-trained models against out-of-distribution (OOD) ImageNet variants. Finally, \autoref{sec:ablation} presents comprehensive ablation studies of our method.

\subsection{Experiment Settings}
\label{sec:exp:settings}
To ensure the efficacy of our methods and design components, we conducted a series of extensive experiments on image classification using ImageNet-1K\cite{ImageNet-1K}, object detection on COCO~\cite{coco} and semantic segmentation on ADE20K\cite{ADE20K}. Unless otherwise stated, our default settings are based on ViT-B.

\paragraph{ImageNet-1K~\cite{ImageNet-1K}} The ImageNet-1K dataset comprises 1.3 million images belonging to 1,000 categories and is divided into training and validation sets. To ensure the fairness of our experiments while applying our methods to MAE~\cite{MAE} and PixMIM\cite{pixmim}, we strictly follow their original pre-training and evaluation settings on ImageNet-1K. This includes following the pre-training schedule, network architecture, learning rate setup, and fine-tuning protocols. Furthermore, in addition to the conventional fine-tuning protocol, we fine-tune the model using a low-shot setting, where only a fraction of the training set (\eg 1\% and 10\%) is used. This approach is consistent with previous works\cite{simclr} and we ensure that the low-shot fine-tuning setting also strictly follows that of the conventional fine-tuning.

\paragraph{ADE20K~\cite{ADE20K}} To conduct the semantic segmentation experiments on ADE20K, we utilize the off-the-shelf settings from MAE\cite{MAE}. With this approach, we fine-tune a UperNet\cite{upernet} for 160k iterations with a batch size of 16 and initialize the relative position bias to zero. 

\paragraph{COCO~\cite{coco}} For our object detection experiments on COCO, we adopt the Mask R-CNN approach\cite{maskrcnn} that enables simultaneous production of bounding boxes and instance masks, with the ViT serving as the backbone. As in MAE, we evaluate the box and mask AP as the metrics for this task. However, we note that there is no universal agreement for the setting of object detection fine-tuning epochs. We have chosen the commonly used 2$\times$ setting, which fine-tunes the model for 25 epochs. Other settings strictly follow those in ViTDet~\cite{ViTDet}.

\paragraph{Ablation studies} We conduct all of our ablation studies based on the customary MAE settings~\cite{MixMIM, CAE}. We pre-train all model variants on ImageNet-1K for 300 epochs and conduct a comprehensive performance comparison on linear probing, fine-tuning, and semantic segmentation. All other settings are consistent with those discussed previously.
\subsection{Main Results}
\label{sec:main_results}
\begin{table*}[!ht]
\centering
\tabcolsep 7pt
\begin{tabular}{lcclllll}
 \multicolumn{3}{c}{Evaluation Protocol$\rightarrow$} & \multicolumn{2}{c}{ImageNet} & \multicolumn{2}{c}{Low-shot} & ADE20K \\
\cmidrule(lr){4-5}\cmidrule(lr){6-7}\cmidrule(lr){8-8}  
Method&Target&Epoch& ft(\%) & lin(\%) & 1\% & 10\% & mIOU \\
\midrule
\multicolumn{8}{@{\;}l}{\bf Supervised learning} \\
\quad DeiT III\cite{DEiT-v3} &- & 800 & 83.8 &- & - & - & 49.3 \\
\midrule
\multicolumn{7}{@{\;}l}{\bf Masked Image Modeling w/ pre-trained target generator} \\
\quad BEiT\cite{BEiT} &DALLE&800&83.2 & 56.7 & - & - & 45.6 \\
\quad CAE\cite{CAE} & DALLE&800&83.8&68.6& - & - & 49.7 \\
\quad MILAN$^{*}$\cite{MILAN} & CLIP-B&400&85.4&78.9&67.5 & 79.7 & 52.7\\
\quad BEiT-v2\cite{BEiTv2}& VQ-KD&1600&85.5& 80.1 &-&-&53.1\\
\quad MaskDistill\cite{MASKdistill} &CLIP-B&800&85.5&-&-&-&54.3\\
\midrule
\multicolumn{7}{@{\;}l}{\bf Masked Image Modeling w/o pre-trained target generator} \\
\quad MaskFeat$^{*}$\cite{MaskFeat}&HOG&1600&84.0&62.3&52.9&73.5 & 48.3\\
\quad SemMAE\cite{SemMAE} & RGB &800&83.4&65.0&- & - & 46.3\\
\quad SimMIM\cite{SimMIM}& RGB &800&83.8&56.7&-&- & -\\
\hdashline
\quad MAE$^{*}$\cite{MAE} & RGB & 300 & 82.8 & 61.5 & 41.4 & 70.5 & 43.9 \\
\quad \textbf{MFF}$_\text{\tt MAE}$ &RGB&300&{{83.3} \more{(+0.5)}}&{63.3} \more{(+1.8)} & 43.7 \more{(+2.3)} & 71.4 \more{(+0.9)} &{47.7} \more{(+3.6)}\\
\quad MAE$^{*}$\cite{MAE} & RGB & 800 & 83.3 & 65.6 & 45.4 & 71.2 & 46.1 \\
\quad \textbf{MFF}$_\text{\tt MAE}$ &RGB&800& 83.6 \more{(+0.3)} & 67.0 \more{(+1.4)} & 48.0 \more{(+2.6)}& 72.0 \more{(+0.8)}  & 47.9 \more{(+1.8)}\\
% \quad MAE$^{*}$\cite{MAE} & RGB & 1600 & 83.5 & 67.8 & 47.8 & 72.4 & 48.1 \\
% \quad \textbf{MFF}$_\text{\tt MAE}$ &RGB&1600& 83.7 \more{(+0.2)} & 69.6 \more{(+1.8)} & 51.9 \more{(+3.1)} & 73.4 \more{(+1.0)}  & 48.3 \more{(+0.2)}\\
\hdashline

\quad PixMIM\cite{pixmim} & RGB&800 & 83.5 & 67.2 & 47.9 & 72.2 & 47.3\\
\quad \textbf{MFF}$_\text{\tt PixMIM}$ &RGB&800&83.6 \more{(+0.1)}&68.2 \more{(+1.0)} & 49.0 \more{(+1.1)} & 73.0 \more{(+0.8)}  &48.6 \more{(+1.3)}\\
% \quad PixMIM\cite{pixmim} & RGB & 1600 & 83.6 & 69.3 & 50.9 & 72.9 & 48.7 \\
% \quad \textbf{MFF}$_\text{\tt PixMIM}$ &RGB&1600&83.9 \more{(+0.3)}&71.1 \more{(+1.8)}& 53.9 \more{(+3.0)} & 74.1 \more{(+1.2)}  &49.1 \more{(+0.4)}\\

\end{tabular}
\caption{\textbf{Performance comparison of MIM methods on various downstream tasks.} We report the results with fine-tuning (ft) and linear probing (lin) experiments on ImageNet-1K, objection detection on COCO, and semantic segmentation on ADE20K. The backbone of all experiments is ViT-B\cite{ViT}. $*$: numbers are reported by running the official code release. Low-shot: end-to-end fine-tuning with 1\% and 10\% of the training set.}
\label{tab:comparison}
\end{table*}

The application of multi-level feature fusion to MAE \cite{MAE} and PixMIM \cite{pixmim} resulted in significant improvements in various downstream tasks, as shown in \autoref{tab:comparison}. After pre-training the model for 300 epochs, we achieve a 0.5\%, 1.8\%, and 3.6\% improvement over MAE in fine-tuning, linear probing, and semantic segmentation, respectively. Additionally, our model exhibits scalability across pre-training epochs and consistently outperforms these base methods by a substantial margin. Compared to these methods, using an extra heavy tokenizer, \eg CLIP, we also gradually close the performance gap with them. Although fine-tuning accuracy is often considered a reliable measure of the quality of non-linear features in a model, we find that it is not a sensitive metric, as compared to other metrics presented in \autoref{tab:comparison}. This may be attributed to pre-training and fine-tuning following the same data distribution, and the size of the training set and model capacity being sufficient to offset the performance gap between different methods. To address this limitation, we adopt the following two workarounds:
% \begin{figure*}[htbp]
% \centering
% \includegraphics[width=\linewidth]{pdf/pic-vit-s.pdf}
% \caption{\textbf{Performance on ViT-S.} Applying \ourmethod to ViT-S brings significant improvements on all downstream tasks. We reuse the same setting as for ViT-B, without specifically tuning.}
% \label{fig:vit-s}
% \end{figure*}
\begin{figure}[htbp]
\centering
\includegraphics[width=\linewidth]{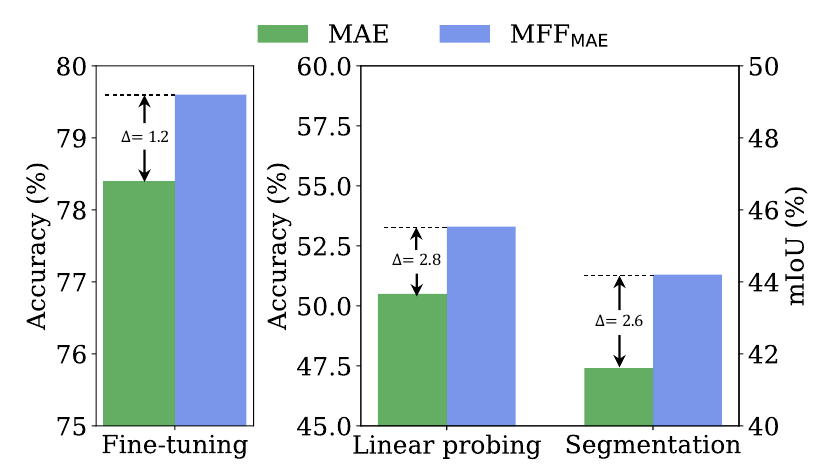}
\caption{\textbf{Performance on ViT-S.} Applying \ourmethod to ViT-S brings significant improvements on all downstream tasks. We reuse the same setting as for ViT-B, without specifically tuning.}
\label{fig:vit-s}
\end{figure}
\paragraph{Low-shot fine-tuning.} This protocol has also been adopted by many previous works, \eg \cite{simclr}. Rather than utilizing the entire training set, we fine-tune the pre-trained model end-to-end using only 1\% and 10\% of the training set. As indicated by \autoref{tab:comparison}, the performance gap between MFF and the base methods is much more prominent when using low-shot fine-tuning, which further verifies the effectiveness of MFF.

\paragraph{Pre-train with ViT-S.} To mitigate the influence brought by model capacity, we pre-train MAE using ViT-S and compare their performance using fine-tuning, linear probing, and semantic segmentation. Since our objective is to evaluate the improvement brought by MFF to these base methods, we do not specifically tune hyper-parameters for the experiments with ViT-S to achieve state-of-the-art performance, but rather use the same settings as for ViT-B. Due to its smaller capacity compared to ViT-B, ViT-S requires a pre-training method that can effectively capture semantic features to perform well on downstream tasks. As demonstrated in \autoref{fig:vit-s}, the method with MFF significantly outperforms their base method, further validating the effectiveness of MFF.

We also evaluate our pre-trained models with the object detection protocol and report the AP$^\text{box}$ and AP$^\text{mask}$. As shown in \autoref{tab:detection}, \ourmethod can still bring non-trivial improvements for object detection.
\begin{table}[ht]
\centering
\tabcolsep 1.5pt
\begin{tabular}{lll}
Method&AP$^\text{box}$&AP$^\text{mask}$\\
\shline
MAE&47.3&42.4\\
\ourmethod$_\text{MAE}$ & 48.0 \more{(+0.7)}& 43.0 \more{(+0.6)} \\
PixMIM & 47.8 & 42.8 \\
\ourmethod$_\text{PixMIM}$ & 48.1 \more{(+0.3)}& 43.1 \more{(+0.3)} \\
\end{tabular}
\vspace{-0.5em}
\caption{\textbf{Results of COCO object detection.}}
\label{tab:detection}
\end{table}

\subsection{Robustness Evaluation}
\label{sec:robust_eval}
Robustness evaluation is a common practice in many previous works \cite{zhou2021ibot,MILAN,MAE} to assess a model's ability to handle different types of noise. In this study, we compare our pre-trained models with their corresponding baselines on four out-of-distribution (OOD) ImageNet variants: ImageNet-Corruption \cite{ImageNetC}, ImageNet-Adversarial \cite{ImageNetA}, ImageNet-Rendition \cite{ImageNetR}, and ImageNet-Sketch \cite{ImageNetS}. These datasets introduce various domain shifts to the original ImageNet-1K and are widely used to evaluate a model's robustness and generalization ability. As illustrated in \autoref{tab:ood}, \ourmethod significantly improves the robustness of MAE and PixMIM on all datasets by a clear margin. The enhanced robustness against domain shifts strengthens the value of our simple yet effective method.

\begin{table*}[ht]
\vspace{-.2em}
\centering
%#################################################
% shallow layer
%#################################################
\subfloat[
\textbf{Which layer for fusion}
\label{tab:shallow_layer}
]{
\begin{minipage}{0.2\linewidth}{
\begin{center}
\tablestyle{4pt}{1.05}
\begin{tabular}{z{30}x{15}x{15}x{15}}
layers &ft&lin&seg\\
\shline
output & 82.8 & 61.5 & 43.9 \\
+shallow & \default{82.9} & \default{61.9} & \default{46.4} \\
+deep & 82.8 & 61.7 & 44.5 \\
% 1,11 & \textbf{82.9} & 61.9 & \textbf{46.7} \\
% 2,11 & 82.9 & \textbf{62.3} & 46.1 \\
% 9,11 & 82.8 & 61.6 & 44.6 \\
% 10,11 & 82.8 & 61.7 & 44.5 \\
% \multicolumn{4}{c}{~}\\
\multicolumn{4}{c}{~}\\
\multicolumn{4}{c}{~}\\
\end{tabular}
\end{center}}\end{minipage}
}
\hspace{1.0em}
%#################################################
% number of layers
%#################################################
\subfloat[
\textbf{Number of layers used for fusion}
\label{tab:layers}
]{
\centering
\begin{minipage}{0.3\linewidth}{
\begin{center}
\tablestyle{4pt}{1.05}
\begin{tabular}{x{40}x{15}x{15}x{15}}
num layers&ft&lin&seg\\
\shline
1 & 82.8 & 61.5 & 43.9 \\
2 & 82.9 & 61.9 & 46.4  \\
% 3 & 82.9 & 62.6 & 46.6 \\
4 & 83.0 & 63.0 & 46.6 \\
6 & \default{83.0} & \default{63.3} & \default{47.7} \\
12 & 83.3 & 62.4 & 46.8 \\
% 1,11 & 82.9 & 62.3 & 46.5  \\
% 1,6,11 & 82.9 & 62.6 & 46.6 \\
% 1,5,8,11 & 83.0 & 63.0 & 46.6 \\
% 1,3,5,7,9,11 & \default{83.0} & \default{63.3} & \default{47.7} \\
% all-layers & 83.3 & 62.4 & 46.8 \\
\end{tabular}
\end{center}}\end{minipage}
}
\hspace{0.5em}
%#################################################
% projection layer
%#################################################
\subfloat[
\textbf{Projection and fusion layer}
\label{tab:proj}
]{
\begin{minipage}{0.4\linewidth}{\begin{center}
\tablestyle{1pt}{1.05}
\begin{tabular}{x{24}x{35}x{24}x{24}x{24}x{24}x{24}}
linear&nonlinear&pool&attn&ft&lin&seg\\
\shline
 & & \checkmark & & 83.0 & 62.7 & 47.2 \\
\checkmark & & \checkmark & & \default{83.0} & \default{63.3} & \default{47.7} \\
 & \checkmark & \checkmark & & 83.0 & 63.0 & 46.2 \\
 \checkmark &  & & \checkmark & 83.1 & 63.0 & 47.3 \\
% \multicolumn{7}{c}{~}\\
\multicolumn{7}{c}{~}\\
\end{tabular}
\end{center}}
\end{minipage}
}
\\
\centering
\vspace{.3em}
%#################################################
\vspace{-.1em}
\caption{We conducted ablation studies with ViT-B/16 that was pre-trained on ImageNet-1K for 300 epochs. Our report includes results for fine-tuning (ft), linear probing (lin), and semantic segmentation (seg). The final settings are highlighted in \colorbox{baselinecolor}{gray}. \textbf{(a)}: \textbf{output} denotes the output layer of the encoder. \textbf{(b)}: the second row corresponds to the second row of (a). \textbf{(c)}: \textbf{linear} and \textbf{nonlinear} denotes linear and nonlinear projection layers, while \textbf{pool} and \textbf{attn} represents the fusion strategies of weight-average pooling and self-attention, respectively. The first row of column (c) does not use any projection layer before fusion.
}
\label{tab:ablations} 
\vspace{-1em}
\end{table*}

\subsection{Ablation Studies}
\label{sec:ablation}
\paragraph{Is shallow layer important?} In order to determine the significance of low-level features from shallow layers, we explore the fusion of the output layer with either a deep or shallow layer. So we try to fuse the output layer with an extra shallow or deep layer, selected from the previous 11 layers of ViT-B\cite{ViT}. And the specific index of the selected layer will be detailed in the appendix. As illustrated in \autoref{tab:shallow_layer}, fusing the output layer with a deep layer only results in marginal improvements. However, incorporating low-level features directly from the shallow layer into the output layer leads to a significant performance boost, as it enables the model to focus on semantic information. Therefore, we have decided to use a shallow layer (\ie the first layer) for multi-level feature fusion.

\paragraph{How many layers are used for fusion ?} Aside from the output layer and the shallow layer picked in the previous selection, it is reasonable to consider using intermediate layers for fusion, as they may contain additional low-level features or high-level meanings that could assist in the reconstruction task. However, selecting these intermediate layers is a daunting task due to the large search space involved. To simplify the process, we pick an additional 1, 2, and 5 layers evenly spaced between the shallow layer and output layer selected in \autoref{tab:shallow_layer}. The specific indices for these selected layers are placed in the appendix. As shown in \autoref{tab:layers}, introducing more layers brings consistent improvements because they may contain unique features, such as textures or colors, that help the model complete the reconstruction task. Nevertheless, when we fuse all these layers, we witness a performance drop in all downstream tasks. This drop may result from the difficulty of optimization, because of the redundancy in these layers.

\begin{table}[h]
\centering
% \scalebox{1}{
\tabcolsep 1.5pt
\begin{tabular}{lllll}
% \toprule 
Method&IN-C$\downarrow$ &IN-A&IN-R&IN-S\\
% \midrule
\shline
MAE&51.7&35.9&48.3&34.5\\
MFF$_\text{\tt MAE}$&49.0 \more{(-2.7)}&37.2 \more{(+1.3)}&51.0 \more{(+2.7)}&36.8 \more{(+2.3)}\\
\hdashline
PixMIM&49.9 &37.1 &49.6 &35.9 \\
MFF$_\text{PixMIM}$&48.5 \more{(-1.4)} &40.1 \more{(+3.0)} &51.6 \more{(+2.0)} &37.8 \more{(+1.9)} \\
\end{tabular}
\vspace{-0.5em}
\caption{\textbf{Robustness evaluation on ImageNet variants.} To evaluate the robustness of MFF, we further evaluate the models (after fine-tuning) from \autoref{tab:comparison} on four ImageNet variants. Results are reported in top-1 accuracy, except for IN-C\cite{ImageNetC} that uses the mean corruption error.}
\label{tab:ood}
\end{table}

\paragraph{Do the projection layer and fusion strategy matter?} In \autoref{eq:proj}, we investigate the influence of the projection layer on the final results. Our findings indicate that a simple linear projection layer is sufficient to achieve satisfactory results, as compared to using no projection layer or a nonlinear projection layer. Incorporating a single linear projection layer offers benefits in mitigating the domain or distribution gap between different layers, as compared to using no projection layer. However, the addition of a nonlinear projection layer, which includes an extra linear projection and GELU activation before the linear projection, introduces computational overhead and is more challenging to optimize. As a result, the non-linear projection achieves sub-optimal performance. With regard to the fusion strategy, we found that the \textbf{weight-average pooling} strategy, which assigns a dynamic weight to each layer and then performs element-wise addition, achieves the best performance. Compared to \textbf{attn}, this strategy shares the merits of simplicity and smaller computational overhead.

\section{Conclusion}
In this study, we take the first step to systematically explore multi-level feature fusion for the isotropic architecture, such as ViT, in masked image modeling. Initially, we recognize that pixel-based MIM approaches tend to excessively rely on low-level features from shallow layers to complete the pixel value reconstruction task by a pilot experiment. We then apply a simple and intuitive multi-level feature fusion to two pixel-based MIM approaches, MAE and PixMIM, and observe significant improvements in both, gradually closing the performance gap with these approaches by using an extra heavy tokenizer. Finally, we conduct an extensive analysis of multi-level feature fusion and find that it can suppress high-frequency information and flatten the loss landscape. We believe that this work can provide the community with a fresh perspective on these pixel-based MIM approaches and continue to rejuvenate this kind of simple and efficient self-supervised learning paradigm.

{\small
\bibliographystyle{ieee_fullname}
\bibliography{egbib}

\begin{thebibliography}{10}\itemsep=-1pt

\bibitem{BEiT}
Hangbo Bao, Li Dong, and Furu Wei.
\newblock {BEiT}: {BERT} pre-training of image transformers.
\newblock {\em ArXiv}, 2021.

\bibitem{dino}
Mathilde Caron, Hugo Touvron, Ishan Misra, Herv\'e J\'egou, Julien Mairal,
  Piotr Bojanowski, and Armand Joulin.
\newblock Emerging properties in self-supervised vision transformers.
\newblock In {\em Proceedings of the IEEE/CVF International Conference on
  Computer Vision (ICCV)}, pages 9650--9660, October 2021.

\bibitem{2021LearningTBP}
Jiacheng Chen, Hexiang Hu, Hao Wu, Yuning Jiang, and Changhu Wang.
\newblock Learning the best pooling strategy for visual semantic embedding.
\newblock {\em IEEE/CVF Conference on Computer Vision and Pattern Recognition
  (CVPR)}, 2021.

\bibitem{pixel}
Mark Chen, Alec Radford, Rewon Child, Jeffrey Wu, Heewoo Jun, David Luan, and
  Ilya Sutskever.
\newblock Generative pretraining from pixels.
\newblock In Hal~Daumé III and Aarti Singh, editors, {\em Proceedings of the
  37th International Conference on Machine Learning}, volume 119 of {\em
  Proceedings of Machine Learning Research}, pages 1691--1703. PMLR, 13--18 Jul
  2020.

\bibitem{simclr}
Ting Chen, Simon Kornblith, Mohammad Norouzi, and Geoffrey Hinton.
\newblock A simple framework for contrastive learning of visual
  representations.
\newblock In {\em International Conference on Machine Learning (ICML)}, 2020.

\bibitem{CAE}
Xiaokang Chen, Mingyu Ding, Xiaodi Wang, Ying Xin, Shentong Mo, Yunhao Wang,
  Shumin Han, Ping Luo, Gang Zeng, and Jingdong Wang.
\newblock Context autoencoder for self-supervised representation learning.
\newblock {\em ArXiv}, abs/2202.03026, 2022.

\bibitem{electra}
Kevin Clark, Minh-Thang Luong, Quoc~V. Le, and Christopher~D. Manning.
\newblock Electra: Pre-training text encoders as discriminators rather than
  generators.
\newblock In {\em International Conference on Learning Representations}, 2020.

\bibitem{ImageNet-1K}
Jia Deng, Wei Dong, Richard Socher, Li-Jia Li, K. Li, and Li Fei-Fei.
\newblock Imagenet: A large-scale hierarchical image database.
\newblock {\em IEEE Conference on Computer Vision and Pattern Recognition
  (CVPR)}, 2009.

\bibitem{ViT}
Alexey Dosovitskiy, Lucas Beyer, Alexander Kolesnikov, Dirk Weissenborn,
  Xiaohua Zhai, Thomas Unterthiner, Mostafa Dehghani, Matthias Minderer, Georg
  Heigold, Sylvain Gelly, Jakob Uszkoreit, and Neil Houlsby.
\newblock An image is worth 16x16 words: Transformers for image recognition at
  scale.
\newblock In {\em International Conference on Learning Representations (ICLR)},
  2021.

\bibitem{EVA}
Yuxin Fang, Wen Wang, Binhui Xie, Quan Sun, Ledell Wu, Xinggang Wang, Tiejun
  Huang, Xinlong Wang, and Yue Cao.
\newblock Eva: Exploring the limits of masked visual representation learning at
  scale.
\newblock {\em arXiv preprint arXiv:2211.07636}, 2022.

\bibitem{ConvMAE}
Peng Gao, Teli Ma, Hongsheng Li, Ziyi Lin, Jifeng Dai, and Yu Qiao.
\newblock {MCMAE}: Masked convolution meets masked autoencoders.
\newblock In {\em Advances in Neural Information Processing Systems (NeurIPS)},
  2022.

\bibitem{ghorbani2019investigation}
Behrooz Ghorbani, Shankar Krishnan, and Ying Xiao.
\newblock An investigation into neural net optimization via hessian eigenvalue
  density.
\newblock In {\em International Conference on Machine Learning}, pages
  2232--2241. PMLR, 2019.

\bibitem{warmup}
Priya Goyal, Piotr Doll{\'a}r, Ross~B. Girshick, Pieter Noordhuis, Lukasz
  Wesolowski, Aapo Kyrola, Andrew Tulloch, Yangqing Jia, and Kaiming He.
\newblock Accurate, large minibatch sgd: Training imagenet in 1 hour.
\newblock {\em ArXiv}, abs/1706.02677, 2017.

\bibitem{MAE}
Kaiming He, Xinlei Chen, Saining Xie, Yanghao Li, Piotr Doll\'ar, and Ross
  Girshick.
\newblock Masked autoencoders are scalable vision learners.
\newblock In {\em Proceedings of the IEEE/CVF Conference on Computer Vision and
  Pattern Recognition (CVPR)}, 2022.

\bibitem{maskrcnn}
Kaiming He, Georgia Gkioxari, Piotr Doll{\'a}r, and Ross Girshick.
\newblock Mask r-cnn.
\newblock In {\em Proceedings of the IEEE international conference on computer
  vision (ICCV)}, 2017.

\bibitem{ImageNetR}
Dan Hendrycks, Steven Basart, Norman Mu, Saurav Kadavath, Frank Wang, Evan
  Dorundo, Rahul Desai, Tyler~Lixuan Zhu, Samyak Parajuli, Mike Guo,
  Dawn~Xiaodong Song, Jacob Steinhardt, and Justin Gilmer.
\newblock The many faces of robustness: A critical analysis of
  out-of-distribution generalization.
\newblock {\em 2021 IEEE/CVF International Conference on Computer Vision
  (ICCV)}, 2021.

\bibitem{ImageNetC}
Dan Hendrycks and Thomas~G. Dietterich.
\newblock Benchmarking neural network robustness to common corruptions and
  perturbations.
\newblock {\em ArXiv}, abs/1903.12261, 2019.

\bibitem{ImageNetA}
Dan Hendrycks, Kevin Zhao, Steven Basart, Jacob Steinhardt, and Dawn~Xiaodong
  Song.
\newblock Natural adversarial examples.
\newblock {\em 2021 IEEE/CVF Conference on Computer Vision and Pattern
  Recognition (CVPR)}, 2021.

\bibitem{randaug}
Elad Hoffer, Tal Ben-Nun, Itay Hubara, Niv Giladi, Torsten Hoefler, and Daniel
  Soudry.
\newblock Augment your batch: better training with larger batches.
\newblock {\em ArXiv}, abs/1901.09335, 2019.

\bibitem{MILAN}
Zejiang Hou, Fei Sun, Yen-Kuang Chen, Yuan Xie, and S.~Y. Kung.
\newblock Milan: Masked image pretraining on language assisted representation.
\newblock {\em ArXiv}, abs/2208.06049, 2022.

\bibitem{hu2021contrast}
Kai Hu, Jie Shao, Yuan Liu, Bhiksha Raj, Marios Savvides, and Zhiqiang Shen.
\newblock Contrast and order representations for video self-supervised
  learning.
\newblock In {\em Proceedings of the IEEE/CVF International Conference on
  Computer Vision (ICCV)}, 2021.

\bibitem{droppath}
Gao Huang, Yu Sun, Zhuang Liu, Daniel Sedra, and Kilian~Q. Weinberger.
\newblock Deep networks with stochastic depth.
\newblock In Bastian Leibe, Jiri Matas, Nicu Sebe, and Max Welling, editors,
  {\em Computer Vision -- ECCV 2016}, pages 646--661, Cham, 2016. Springer
  International Publishing.

\bibitem{CMAE}
Zhicheng Huang, Xiaojie Jin, Cheng Lu, Qibin Hou, Mingg-Ming Cheng, Dongmei Fu,
  Xiaohui Shen, and Jiashi Feng.
\newblock Contrastive masked autoencoders are stronger vision learners.
\newblock {\em ArXiv}, abs/2207.13532, 2022.

\bibitem{SemMAE}
Gang Li, Heliang Zheng, Daqing Liu, Chaoyue Wang, Bing Su, and Changwen Zheng.
\newblock Sem{MAE}: Semantic-guided masking for learning masked autoencoders.
\newblock In {\em Advances in Neural Information Processing Systems (NeurIPS)},
  2022.

\bibitem{ViTDet}
Yanghao Li, Hanzi Mao, Ross Girshick, and Kaiming He.
\newblock Exploring plain vision transformer backbones for object detection.
\newblock In {\em ECCV}, 2022.

\bibitem{FPN}
Tsung-Yi Lin, Piotr Doll{\'a}r, Ross Girshick, Kaiming He, Bharath Hariharan,
  and Serge Belongie.
\newblock Feature pyramid networks for object detection.
\newblock In {\em Proceedings of the IEEE conference on computer vision and
  pattern recognition}, pages 2117--2125, 2017.

\bibitem{coco}
Tsung-Yi Lin, Michael Maire, Serge Belongie, James Hays, Pietro Perona, Deva
  Ramanan, Piotr Doll{\'a}r, and C~Lawrence Zitnick.
\newblock Microsoft coco: Common objects in context.
\newblock In {\em ECCV}, 2014.

\bibitem{MixMIM}
Jihao Liu, Xin Huang, Yu Liu, and Hongsheng Li.
\newblock Mixmim: Mixed and masked image modeling for efficient visual
  representation learning.
\newblock {\em arXiv preprint arXiv:2205.13137}, 2022.

\bibitem{2022MoQuad}
Yuan Liu, Jiacheng Chen, and Hao Wu.
\newblock Moquad: Motion-focused quadruple construction for video contrastive
  learning.
\newblock {\em ArXiv}, abs/2212.10870, 2022.

\bibitem{pixmim}
Yuan Liu, Songyang Zhang, Jiacheng Chen, Kai Chen, and Dahua Lin.
\newblock Pixmim: Rethinking pixel reconstruction in masked image modeling.
\newblock {\em arXiv preprint arXiv:2303.02416}, 2023.

\bibitem{cosine}
Ilya Loshchilov and Frank Hutter.
\newblock {SGDR}: Stochastic gradient descent with warm restarts.
\newblock In {\em International Conference on Learning Representations}, 2017.

\bibitem{adamw}
Ilya Loshchilov and Frank Hutter.
\newblock Decoupled weight decay regularization.
\newblock In {\em International Conference on Learning Representations}, 2019.

\bibitem{howdo}
Namuk Park and Songkuk Kim.
\newblock How do vision transformers work?
\newblock In {\em International Conference on Learning Representations}, 2022.

\bibitem{BEiTv2}
Zhiliang Peng, Li Dong, Hangbo Bao, Qixiang Ye, and Furu Wei.
\newblock Beit v2: Masked image modeling with vector-quantized visual
  tokenizers.
\newblock {\em ArXiv}, abs/2208.06366, 2022.

\bibitem{MASKdistill}
Zhiliang Peng, Li Dong, Hangbo Bao, Qixiang Ye, and Furu Wei.
\newblock A unified view of masked image modeling.
\newblock {\em ArXiv}, abs/2210.10615, 2022.

\bibitem{CLIP}
Alec Radford, Jong~Wook Kim, Chris Hallacy, Aditya Ramesh, Gabriel Goh,
  Sandhini Agarwal, Girish Sastry, Amanda Askell, Pamela Mishkin, Jack Clark,
  Gretchen Krueger, and Ilya Sutskever.
\newblock Learning transferable visual models from natural language
  supervision.
\newblock In {\em ICML}, 2021.

\bibitem{DALLE}
Aditya Ramesh, Mikhail Pavlov, Gabriel Goh, Scott Gray, Chelsea Voss, Alec
  Radford, Mark Chen, and Ilya Sutskever.
\newblock Zero-shot text-to-image generation.
\newblock {\em ArXiv}, abs/2102.12092, 2021.

\bibitem{ren2023deepmim}
Sucheng Ren, Fangyun Wei, Samuel Albanie, Zheng Zhang, and Han Hu.
\newblock Deepmim: Deep supervision for masked image modeling.
\newblock {\em arXiv preprint arXiv:2303.08817}, 2023.

\bibitem{smooth}
Christian Szegedy, Vincent Vanhoucke, Sergey Ioffe, Jon Shlens, and Zbigniew
  Wojna.
\newblock Rethinking the inception architecture for computer vision.
\newblock In {\em 2016 IEEE Conference on Computer Vision and Pattern
  Recognition (CVPR)}, pages 2818--2826, 2016.

\bibitem{itpn}
Yunjie Tian, Lingxi Xie, Zhaozhi Wang, Longhui Wei, Xiaopeng Zhang, Jianbin
  Jiao, Yaowei Wang, Qi Tian, and Qixiang Ye.
\newblock Integrally pre-trained transformer pyramid networks.
\newblock {\em arXiv preprint arXiv:2211.12735}, 2022.

\bibitem{DeiT}
Hugo Touvron, Matthieu Cord, Matthijs Douze, Francisco Massa, Alexandre
  Sablayrolles, and Herv'e J'egou.
\newblock Training data-efficient image transformers \& distillation through
  attention.
\newblock In {\em ICML}, 2021.

\bibitem{DeiT-v2}
Hugo Touvron, Matthieu Cord, Alaaeldin El-Nouby, Jakob Verbeek, and Herve
  Jegou.
\newblock Three things everyone should know about vision transformers.
\newblock {\em arXiv preprint arXiv:2203.09795}, 2022.

\bibitem{DEiT-v3}
Hugo Touvron, Matthieu Cord, and Herv{\'e} J{\'e}gou.
\newblock Deit iii: Revenge of the vit.
\newblock In Shai Avidan, Gabriel Brostow, Moustapha Ciss{\'e}, Giovanni~Maria
  Farinella, and Tal Hassner, editors, {\em ECCV}, 2022.

\bibitem{denoising}
Pascal Vincent, Hugo Larochelle, Yoshua Bengio, and Pierre-Antoine Manzagol.
\newblock Extracting and composing robust features with denoising autoencoders.
\newblock In {\em International Conference on Machine Learning (ICML)}, 2008.

\bibitem{ImageNetS}
Haohan Wang, Songwei Ge, Eric~P. Xing, and Zachary~Chase Lipton.
\newblock Learning robust global representations by penalizing local predictive
  power.
\newblock In {\em NeurIPS}, 2019.

\bibitem{MaskFeat}
Chen Wei, Haoqi Fan, Saining Xie, Chaoxia Wu, Alan~Loddon Yuille, and Christoph
  Feichtenhofer.
\newblock Masked feature prediction for self-supervised visual pre-training.
\newblock {\em 2022 IEEE/CVF Conference on Computer Vision and Pattern
  Recognition (CVPR)}, 2022.

\bibitem{upernet}
Tete Xiao, Yingcheng Liu, Bolei Zhou, Yuning Jiang, and Jian Sun.
\newblock Unified perceptual parsing for scene understanding.
\newblock In {\em ECCV}, 2018.

\bibitem{SimMIM}
Zhenda Xie, Zheng Zhang, Yue Cao, Yutong Lin, Jianmin Bao, Zhuliang Yao, Qi
  Dai, and Han Hu.
\newblock Simmim: A simple framework for masked image modeling.
\newblock In {\em International Conference on Computer Vision and Pattern
  Recognition (CVPR)}, 2022.

\bibitem{lars}
Yang You, Igor Gitman, and Boris Ginsburg.
\newblock Large batch training of convolutional networks.
\newblock {\em arXiv: Computer Vision and Pattern Recognition}, 2017.

\bibitem{cutmix}
Sangdoo Yun, Dongyoon Han, Sanghyuk Chun, Seong~Joon Oh, Youngjoon Yoo, and
  Junsuk Choe.
\newblock Cutmix: Regularization strategy to train strong classifiers with
  localizable features.
\newblock In {\em 2019 IEEE/CVF International Conference on Computer Vision
  (ICCV)}, pages 6022--6031, 2019.

\bibitem{mixup}
Hongyi Zhang, Moustapha Cisse, Yann~N. Dauphin, and David Lopez-Paz.
\newblock mixup: Beyond empirical risk minimization.
\newblock In {\em International Conference on Learning Representations}, 2018.

\bibitem{ADE20K}
Bolei Zhou, Hang Zhao, Xavier Puig, Sanja Fidler, Adela Barriuso, and Antonio
  Torralba.
\newblock Semantic understanding of scenes through the ade20k dataset.
\newblock {\em International Journal of Computer Vision (IJCV)}, 2018.

\bibitem{zhou2021ibot}
Jinghao Zhou, Chen Wei, Huiyu Wang, Wei Shen, Cihang Xie, Alan Yuille, and Tao
  Kong.
\newblock ibot: Image bert pre-training with online tokenizer.
\newblock {\em International Conference on Learning Representations (ICLR)},
  2022.

\bibitem{ibot}
Jinghao Zhou, Chen Wei, Huiyu Wang, Wei Shen, Cihang Xie, Alan Yuille, and Tao
  Kong.
\newblock Image {BERT} pre-training with online tokenizer.
\newblock In {\em International Conference on Learning Representations (ICLR},
  2022.

\end{thebibliography}
}

\appendix

\section{Appendix}

\subsection{Pre-training}

The settings for pre-training strictly follows those in MAE\cite{MAE} and PixMIM\cite{pixmim}, with details shown below:
\begin{table}[h]
\tablestyle{6pt}{1.02}
\begin{tabular}{y{96}|y{68}}
config & value \\
\shline
optimizer & AdamW \cite{adamw} \\
base learning rate & 1.5e-4 \\
weight decay & 0.05 \\
optimizer momentum & $\beta_1, \beta_2{=}0.9, 0.95$~\cite{pixel} \\
batch size & 4096 \\
learning rate schedule & cosine decay \cite{cosine} \\
warmup epochs \cite{warmup} & 40 \\
\end{tabular}
\caption{{Pre-training setting of MFF$_\text{MAE}$ and MFF$_\text{PixMIM}$}}
\label{tab:impl_mae_pretrain} 
\end{table}

\subsection{Fine-tuning and linear probing}
We also stick to the settings in MAE\cite{MAE} for the ViT-B\cite{ViT} model concerning fine-tuning and linear probing. Since our objective is to measure the enhancement brought by MFF and not attain the state-of-the-art (SOTA) performance, we employ the same settings as ViT-B without any specific adjustments for ViT-S.

\begin{table}[h]
\tablestyle{6pt}{1.02}
% \scriptsize
\begin{tabular}{y{96}|y{68}}
config & value \\
\shline
optimizer & LARS \cite{lars} \\
base learning rate & 0.1 \\
weight decay & 0 \\
optimizer momentum & 0.9 \\
batch size & 16384 \\
learning rate schedule & cosine decay \\
warmup epochs & 10 \\
training epochs & 90 \\
augmentation & RandomResizedCrop \\
\end{tabular}
% \vspace{-.5em}
\caption{{Linear probing setting of \ourmethod$_\text{MAE}$ and \ourmethod$_\text{PixMIM}$.}}
\label{tab:impl_mae_linear}
\end{table}
\begin{table}[h]
\tablestyle{6pt}{1.02}
% \scriptsize
\begin{tabular}{y{96}|y{68}}
config & value \\
\shline
optimizer & AdamW\cite{adamw} \\
base learning rate & 1e-3 \\
weight decay & 0.05 \\
optimizer momentum & $\beta_1, \beta_2{=}0.9, 0.999$ \\
layer-wise lr decay \cite{electra, BEiT} & 0.75 \\
batch size & 1024 \\
learning rate schedule & cosine decay \\
warmup epochs & 5 \\
training epochs & 100  \\
augmentation & RandAug (9, 0.5) \cite{randaug} \\
label smoothing \cite{smooth} & 0.1 \\
mixup \cite{mixup} & 0.8 \\
cutmix \cite{cutmix} & 1.0 \\
drop path \cite{droppath} & 0.1 \\
\end{tabular}
\caption{{End-to-end fine-tuning setting of \ourmethod$_\text{MAE}$, \ourmethod$_\text{PixMIM}$}}
\label{tab:impl_mae_finetune} 
\end{table}

\subsection{Object detection and segmentation in COCO}
All these settings also strictly follow those in MAE\cite{MAE} but choose the commonly used 2$\times$ settings, which fine-tunes the model on COCO\cite{coco} for 25 epochs.

\subsection{Semantic segmentation in ADE20K}
We stick to the settings used in MAE\cite{MAE} and PixMIM\cite{pixmim}, fine-tuning the pre-trained model end-to-end for 16k iterations with a batch size of 16.

\subsection{Selected indices of the ablation study}
Inspired by the results of the pilot experiment depicted in Figure 1 of the main paper, we choose layer$_0$ as the shallow layer, and layer$_{10}$ as the deep layer for the ablation experiment outlined in Table 3(a). Additionally, for ablation study in Table 3(b), we have selected additional two, four, and ten layers, evenly distributed between layer$_0$ and the output layer (layer$_{11}$). The detailed indices for Table 3(b) is shown in the \autoref{tab:indices}. 

\begin{table}[h]
\centering
\tabcolsep 1.5pt
\begin{tabular}{cc}
num layers & indices \\
\shline
1 & 11 \\
2 & 0,11 \\
4 & 0,4,8,11 \\
6 & 0,2,4,6,8,11 \\ 
12 & 0,1,2,3,4,5,6,7,8,9,10,11 \\
\end{tabular}
\vspace{-0.5em}
\caption{\textbf{Detailed indices for Table 3(b) of the main paper.} We try to make the additionally selected indices are evenly distributed between the first layer and last layer.}
\label{tab:indices}
\end{table}

\begin{figure*}[!htbp]
\centering
\includegraphics[width=\linewidth,scale=1.00]{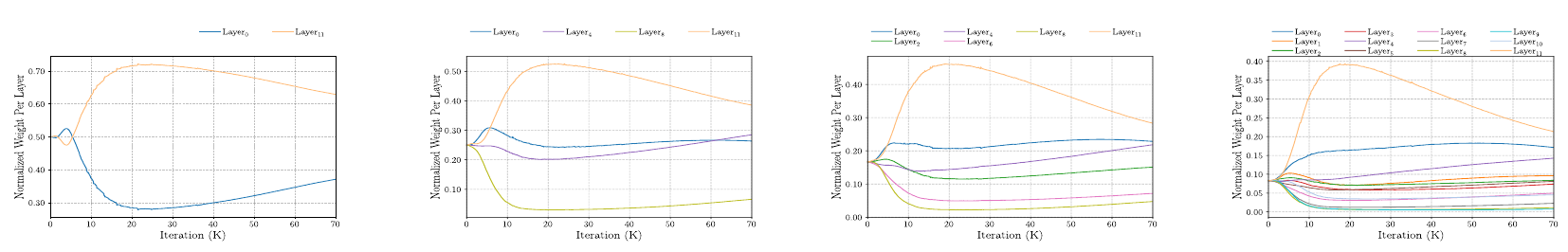}
\caption{\textbf{Model increasingly relies on shallow layers.}}
\label{fig:layer_weights}
\end{figure*}

In addition, similar to the pilot experiment in Figure 1 of the main paper, we observe the weight for each layer of all experiments in Table 3(b) of the main paper. Just as shown in \autoref{fig:layer_weights}, no matter in which case, the model increasing relies on these shallow layers for the reconstruction tasks, indicating the significance of injecting low-level information into the output layer.

\subsection{Transfer learning}
We also study transfer learning where we pre-train on ImageNet-1K and fine-tune on several smaller datasets. We follow the training recipe and protocol in DINO\cite{dino}. \ourmethod$_\text{MAE}$ consistently outperforms MAE on CIFAR10, CIFAR100, and Stanford Cars. As shown in the following table, \ourmethod$_\text{MAE}$ consistently improves MAE on all datasets.

\begin{table}[h]
\centering
% \scalebox{1}{
\tabcolsep 1.5pt
\begin{tabular}{lllll}
Method& Epoch & CIFAR10 & CIFAR100 & Cars\\
\shline
MAE & 800 & 98.4 & 89.4 & 94.3 \\
\ourmethod$_\text{MAE}$ & 800 & 98.6 \more{(+0.2)} & 90.3 \more{(+0.9)} & 94.7 \more{(+0.4)} \\
\end{tabular}
\vspace{-0.5em}
\caption{\textbf{Transfer learning on smaller datasets.}}
\label{tab:transfer}
\end{table}

\subsection{Feature-based MIM does not Suffer from being Biased toward Low-level Feature}
To supplement the findings in \autoref{fig:weight_vs_layer}, we apply multi-level feature fusion (MFF) to EVA\cite{EVA} and MILAN\cite{MILAN}, and evaluate their performance with linear probing, fine-tuning and semantic segmentation. Detailed results are shown below:

\begin{table}[h]
\centering
\tabcolsep 1.5pt
\begin{tabular}{llccc}
Method&Epoch&lin&seg&ft                               \\
\shline
EVA                & 400 & 69.0 & 49.5 & 83.7         \\
MFF$_\text{EVA}$   & 400 & 68.9  & 49.4  & 83.8       \\
MILAN                & 400 & 79.9 & 52.7 & 85.4       \\
MFF$_\text{MILAN}$   & 400 & 79.7  & 52.9  & 85.0     \\
\end{tabular}
\end{table}

\noindent As shown in the table above, MFF brings marginal improvements to feature-based MIMs, consistent with the findings in \autoref{fig:weight_vs_layer}. 

\subsection{The Effect of Deep Supervision}
To exclude the influence of deep supervision\cite{ren2023deepmim}, we detach all shallow layers before fusing with the last layer (MFF$_\text{MAE}^\text{detach}$), ensuring that gradients do not propagate through these shortcuts to the shallow layers. As shown in the table below, deep supervision alone does not improve MAE, and MFF's improvements come from alleviating the problem of being biased toward high-freq components.

\begin{table}[h]
\centering
\small
% \scalebox{1}{
\tabcolsep 1.5pt
\begin{tabular}{llllll}
% \toprule 
Method&model&epoch&lin&seg&ft\\
\shline
MFF$_\text{MAE}$ & ViT-B            & 800 & 67.0 & 47.9 & 83.6              \\
MFF$_\text{MAE}^\text{detach}$ & ViT-B  & 800 & 66.8 \red{(-0.2)} & 48.0 \green{(+0.1)} & 83.5\red{(-0.1)}              \\
\end{tabular}
\vspace{-0.5em}
\label{tab:r-1}
\end{table}

\end{document}